\pdfoutput=1

\documentclass[11pt]{article}

\usepackage[]{ACL2023}
\usepackage{comment}

\usepackage{times}
\usepackage{latexsym}
\usepackage{booktabs}

\usepackage[T1]{fontenc}

\usepackage[utf8]{inputenc}
\usepackage{multirow}

\usepackage{microtype}

\usepackage{graphicx}
\graphicspath{ {figures} }

\usepackage{inconsolata}
\usepackage[acronyms]{glossaries}

\usepackage{pgfplots, pgfplotstable}

\newacronym{pgt}{PGTask}{Profile Generation Task}
\newacronym{pgd}{PGDataset}{Profile Generation Dataset}
\newacronym{nlp}{NLP}{Natural Language Processing}
\newacronym{nli}{NLI}{Natural Language Inference}
\newacronym{dnli}{DNLI}{Dialogue Natural Language Inference}
\newacronym{clm}{CLM}{causal language modeling}

\DeclareMathOperator*{\argmax}{arg\,max}

\DeclareMathOperator*{\softmax}{softmax}

\title{PGTask: Introducing the Task of Profile Generation from Dialogues}

\author{Rui Ribeiro, Joao P. Carvalho, Luísa Coheur\\
  INESC-ID, Lisboa\\
  Instituto Superior Técnico, Universidade de Lisboa \\
  \texttt{\{rui.m.ribeiro, joao.carvalho, luisa.coheur\}@inesc-id.pt} \\}

\begin{document}
\maketitle

\begin{abstract}
Recent approaches have attempted to personalize dialogue systems by leveraging profile information into models. However, this knowledge is scarce and difficult to obtain, which makes the extraction/generation of profile information from dialogues a fundamental asset. To surpass this limitation, we introduce the \gls*{pgt}. We contribute with a new dataset for this problem, comprising profile sentences aligned with related utterances, extracted from a corpus of dialogues. Furthermore, using state-of-the-art methods, we provide a benchmark for profile generation on this novel dataset. Our experiments disclose the challenges of profile generation, and we hope that this introduces a new research direction.
\end{abstract}

\section{Introduction}

Building conversational systems that mimic human attributes has always been a long-term goal in \gls*{nlp}. 
Various works have attempted to leverage speaker profile information to improve the consistency of dialogue generation models \cite{EXPLOIT-PERSONA, LONG-TERM, DATA-MANIPULATION-PERSONA}. 
By incorporating speaker-specific characteristics, such as age, gender, personality traits, and cultural background, into the conversational systems, it is possible to create more personalized and human-like interactions.
However, for dialogue systems, this sort of information is scarce and requires annotation efforts that are expensive to obtain, so there is a need to build methods that automatically gather this knowledge from dialogues.

\citet{PERSONACHAT} introduced PersonaChat, a dataset comprising a collection of profile sentences (\textit{persona}) that reflect each speaker's individual characteristics and personal facts. These profiles serve as a knowledge base for promoting the consistency between utterances from speakers, and various recent dialogue models have incorporated this information using diverse techniques \cite{GENERATE-DELETE-REWRITE, BERTOVERBERT, DATA-MANIPULATION-PERSONA}.

Few works have attempted to infer profile information from PersonaChat dialogues.
\citet{DETECT-SPEAKER} restructured PersonaChat and built the Persona Detection Task, where the goal was to retrieve the correct persona amongst a set of distractor personas. Although introducing an interesting research path, this task is limited to a set of pre-defined personas, which is not suitable for extracting profile sentences from unseen conversational data.
\citet{DATA-MANIPULATION-PERSONA} also manipulate PersonaChat to incorporate model-agnostic personas into the dialogue generation task. Nevertheless, for the profile generation task, PersonaChat is structured in a profile-to-dialogue manner and lacks information about the corresponding profile sentence per turn, which may become a challenge when the task becomes extracting profile information from utterances.

\begin{figure}[t]
    \centering
 \includegraphics[width=0.50\textwidth]{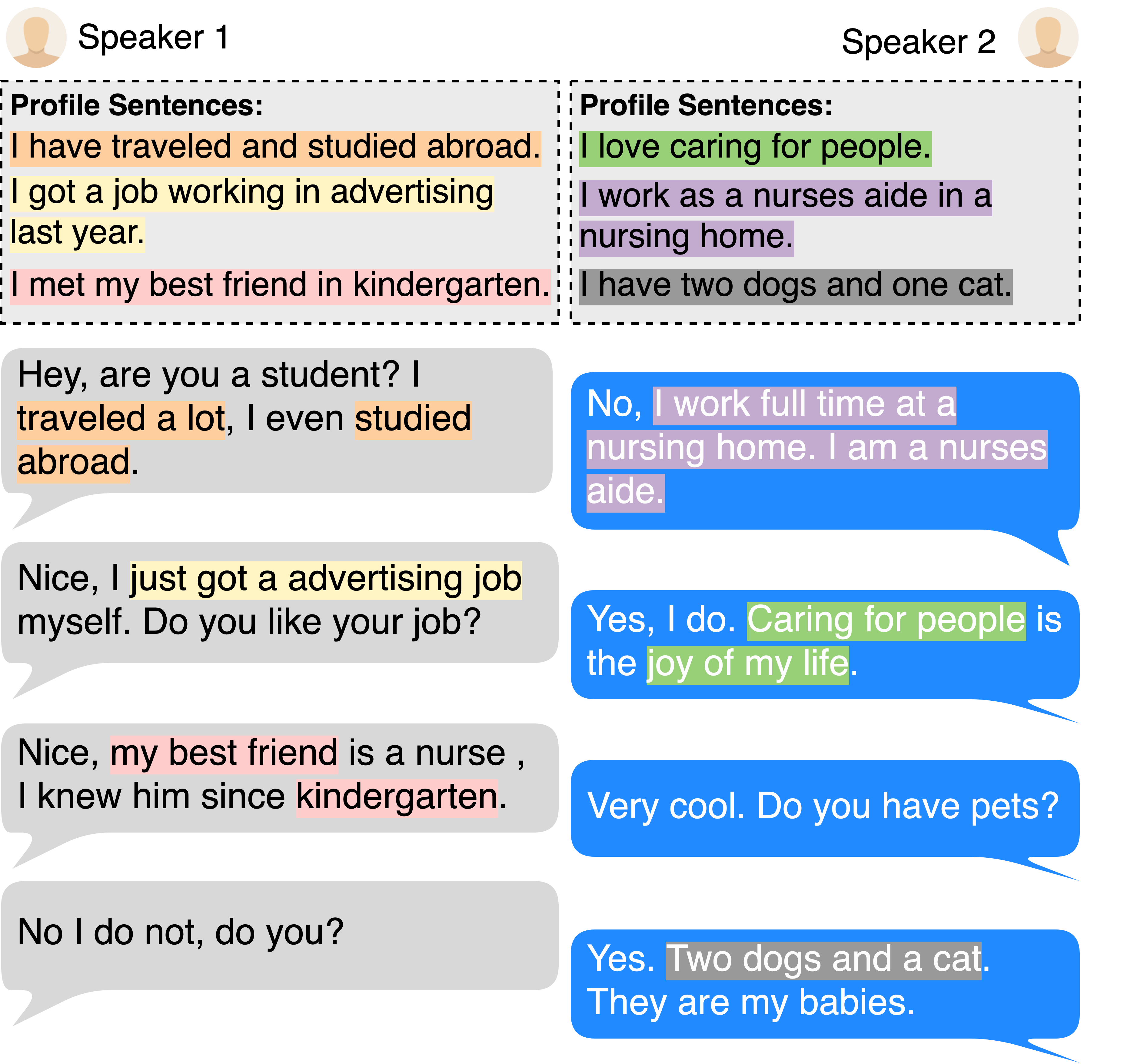}
    \caption{An example dialogue where each turn contains the corresponding profile sentence.}
    \label{fig:example-dialogue}
\end{figure}

In this work, we introduce the \gls*{pgt}\footnote{Dataset and code are available at \href{https://github.com/ruinunca/PGTask}{https://github.com/ruinunca/PGTask}. 
}, where the goal is to generate profile sentences given speaker utterances. For this, we create a new dataset, the \gls*{pgd}, which relates utterances with the respective profile sentences upon the existing PersonaChat corpus. In Figure \ref{fig:example-dialogue}, we can observe several examples of relations between profile sentences and the corresponding speaker's utterance. Notice, however, that the task is more challenging than just finding, within the dialogues, utterances that highly relate to each profile sentence. For instance, the profile sentence ``I like all genres of music.'' is probably at the origin of the utterance ``Yes, sometimes I also listen to classical music.'', but we cannot extract that profile sentence from that single utterance (the goal of \gls*{pgt}).

We framed our problem as an entailment classification task and, after human feedback, we reached the final \gls*{pgd}. Finally, we provide results from three state-of-the-art models trained and evaluated in the proposed dataset. 



\section{Building \gls*{pgd}}

In this section, we demonstrate how we formulated our task as an entailment detection problem and describe the utilization of human experts' feedback to build a consistent dataset.

\subsection{Modeling Entailment Relations}\label{sec:model}

In the \gls*{nli} task, the goal is to classify the relationship between a pair of premise and hypothesis sentences into three classes: entailment (E), neutral (N), and contradiction (C). \citet{DNLI} extended the \gls*{nli} task to the dialogue setting and introduced the \gls*{dnli} dataset, where the input sentences consist of dialogue utterances from PersonaChat.
We adopt this procedure and train a model \(\mathcal{M}^{NLI}\) to identify the correct profile sentences for each utterance in a dialogue.

Consider two sentences \(s_i\) and \(s_j\) that are concatenated into the input \(x= \{s_i, s_j\}\).
First, we utilize RoBERTa \cite{ROBERTA} to obtain a hidden representation \(h\) from the input \(x\).
Then, we include a \(\softmax\) classifier on top of RoBERTa to obtain the probability distribution over the set of possible classes.
Formally, we obtain the probability of label \(y \in \{C, N, E\}\) with:

\begin{equation}
    \begin{split}
    h &= \text{RoBERTa}(x), \\
    p_{\mathcal{M}^{NLI}}(y|x) &= \softmax(Wh),
    \end{split}
\end{equation}

where \(W\) is the learnable parameter matrix from the classification layer. 
We fine-tune both RoBERTa and \(W\) parameters by maximizing the log-probability of the correct label.

\begin{table}[ht]
    \centering
    \begin{tabular}{lc}
    \hline
    \textbf{Datasets} & \textbf{Accuracy (\%)}\\
    \hline
    DNLI & 91.24 \\
    MNLI + DNLI & \textbf{91.75} \\
    \hline
    \end{tabular}
\caption{Accuracy of fine-tuned \textsc{RoBERTa} for the test set of DNLI.}
\label{tab:acc-dnli}
\end{table}

We experiment with two different settings where we fine-tune RoBERTa only on \gls*{dnli} and on MNLI \cite{MNLI}, a benchmark multi-genre NLI dataset, and \gls*{dnli} datasets for better generalization.
Details are provided in Appendix \ref{appendix:roberta-dnli}.
Table \ref{tab:acc-dnli} shows the results on the test set, where the latter achieves higher accuracy and is selected as the annotation model.

\subsection{Dataset Annotation}

In PersonaChat \cite{PERSONACHAT}, each dialogue carries a set of profile sentences for both speakers.
Consider a set of \(n\) utterances from a speaker, \(U=\{u_1, u_2, ..., u_n\}\), a set of \(k\) profile sentences \(P=\{p_1, p_2, ..., p_k\}\) from the same speaker, and the dialogue \gls*{nli} model from Section \ref{sec:model}.
Then, at time step \(t\), we can determine one or more profile sentences \(s_t\) related to utterance \(u_t\) using:


\begin{equation}
\label{eq:build-profile}
    \begin{split}
    s_t =  \{p_i &\in P : \\
    &\argmax_{y \in \{C, N, E\}}(p_{\mathcal{M}^{NLI}}(y|\{u_t, p_i\}) = E \}.
    \end{split}
\end{equation}

In Equation \ref{eq:build-profile}, the profile sentences are gathered by considering the entailed cases between the utterances and the profile sentences, where each utterance could be associated with more than one profile sentence. 
In Table \ref{tab:example}, we provide an extract from the \gls*{pgd}.

\begin{table}[ht]
    \centering
    \begin{tabular}{p{0.45\columnwidth}|p{0.45\columnwidth}}
    \hline
    \textbf{Utterance} & \textbf{Profile Sentences} \\
    \hline 
    I enjoy hanging with my mother she is my best friend. & My mom is my best friend.\\
    \hline
    \multirow{2}{0.45\columnwidth}{I am almost done, I only have two years left in law school.} & I have got two more years in college. \\ \cline{2-2}
    & I study law. \\
    \hline
    \end{tabular}
\caption{Two examples from \gls*{pgd}.}
\label{tab:example}
\end{table}

\subsection{Human Annotations}

In the profile generation task, the profile must represent a possible extraction from the dialogue utterance, and this correlation's direction between the utterance and the profile sentence must be valid.
To assess the quality of the automatic annotations from our model, we resort to human evaluation.

\begin{figure}[h]
\centering
\resizebox {\columnwidth} {!} {
    \begin{tikzpicture}[object/.style={thin,double,<->}]
    \begin{axis}[
        ymin=0, ymax=8570, 
        xmax=100,
        minor y tick num = 2,
        area style,
        ylabel={\# Counts},
        xlabel={\(p_{\mathcal{M}^{NLI}}(E)\) [\%]},
        enlargelimits=false,
        ]
    \addplot table [col sep=semicolon] {points_counts_per.csv} \closedcycle; 
    \end{axis}
    \end{tikzpicture}
}
\vskip-0.2cm
\caption{Distribution of the entailment class probability for the entailed cases (\(\mu = 93.4\), \(\sigma^2 = 1.10\)).}
\label{fig:confidence-plot}
\end{figure}
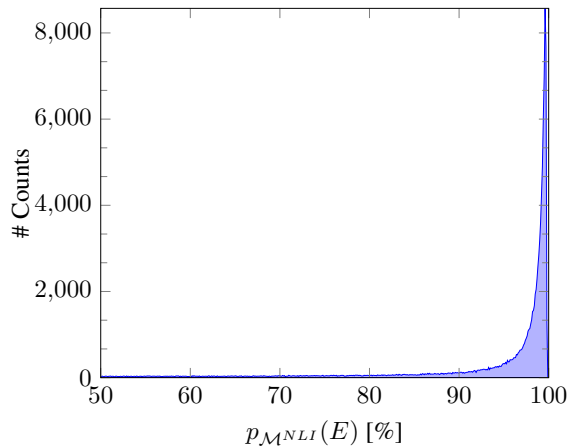

For all the pairs classified as entailed in Equation \ref{eq:build-profile}, we measure the confidence by inspecting the \(\softmax\) probability assigned to the entailment class.
Our intuition is that a weak confidence when classifying a profile sentence as entailed corresponds to a weak or incorrect correlation and can be removed from the dataset.
In Figure \ref{fig:confidence-plot}, we plot the distribution of the scores from the entailment class for all points obtained from Equation \ref{eq:build-profile}.

To determine if a higher confidence value corresponds to a correct example, we randomly select 100 samples from 3 intervals: \([50, 70]\), \(]70, 90]\), and \(]90, 100]\).
We asked 3 expert annotators from our department to ``mark with an X if the profile sentence could be extracted from the given utterance''.
The quality of the samples is measured by the number of marked samples by the annotators (accuracy).
The agreement rate between annotators was 86.66\% and the average accuracy for each interval was 8.33\%, 12.33\%, and 51.67\%, respectively.
The results obtained show that when the confidence of the model grows, the correlation between the profile sentence and the utterance also increases.

After inspecting the results from the annotators, we observed that most of the marked samples had more than 99\% confidence.
We asked for a second round of annotations with 100 samples but now only for samples with more than 99\% confidence.
The agreement rate between annotators was 91\% and the average accuracy was 87,33\%, a significantly higher score compared to the \(]90, 100]\) interval. We decided, thus, that \gls*{pgd} only considers the samples which the model classified with more than 99\% confidence.


\subsection{\gls*{pgd} Statistics}

\begin{table}
\centering
\begin{tabular}{lll}
\hline
\multirow{4}{*}{\textbf{Train}} & \# Samples & 34355 \\
&Avg. Profile Sentences & 1.06 \\
&Avg. Utterance Words & 13.13 \\
&Avg. Profile Sentence Words & 7.14 \\
\hline
\multirow{4}{*}{\textbf{Valid}} & \# Samples & 4236 \\
&Avg. Profile Sentences & 1.06 \\
&Avg. Utterance Words & 13.36 \\
&Avg. Profile Sentence Words & 7.67 \\
\hline
\multirow{4}{*}{\textbf{Test}} & \# Samples & 3760 \\
&Avg. Profile Sentences & 1.06 \\
&Avg. Utterance Words & 13.05 \\
&Avg. Profile Sentence Words & 7.17 \\
\hline
\end{tabular}
\caption{Dataset Statistics.}
\label{tab:dataset-statistics}
\end{table}

In Table \ref{tab:dataset-statistics}, we provide the dataset statistics for the gathered samples.

\section{Benchmarking the \gls*{pgt}}

\begin{table*}[t]
\centering
\small
    \centering
    \begin{tabular}{l|l|cccc|ccc|c}
    \toprule
    \multicolumn{2}{c}{Model} & BLEU-1 & BLEU-2 & BLEU-3 & BLEU-4 & ROUGE-1 & ROUGE-2 & ROUGE-L & BERTScore \\
    \midrule
    \parbox[t]{2mm}{\multirow{3}{*}{\rotatebox[origin=c]{90}{W/o FT}}} &\texttt{distilgpt2}& 5.59 & 0.30 &0.00 &0.00 &6.86 &0.93&5.80 &84.66 \\
    &\texttt{gpt2-small}& 4.87&0.40	&0.00	&0.00	&6.08	&0.63	&5.20	&84.21\\
    &\texttt{gpt2-medium} & 4.48 &0.20 &0.00 &0.00 &7.20 &0.31&5.32	&83.28\\
    \midrule
    \parbox[t]{2mm}{\multirow{3}{*}{\rotatebox[origin=c]{90}{W/ FT}}} &\texttt{distilgpt2}&44.42&13.18	&5.60	&0.00	&35.68	&14.12	&35.39	&92.35 \\
    &\texttt{gpt2-small}&61.30	&32.30	&20.62	&9.44	&50.07	&28.31	&50.00	&94.39\\
    &\texttt{gpt2-medium}&59.31&25.94&15.30&9.17&46.32&24.14	&45.88	&94.76\\
    \bottomrule
    \end{tabular}
\caption{Generation results for models with and without fine-tuning (FT) on the \gls*{pgd}. The results presented are the average score of 5 runs. The scores range between 0 and 100\%.}
\label{tab:gen-results}
\end{table*}

In this task, the goal is to generate a profile sentence conditioned on an utterance.
Transformer-based decoders have achieved substantial progress in various \gls*{nlp} tasks \cite{GPT2}.
We leverage these models and rely on a \gls*{clm} objective for our profile generation task.
More precisely, considering a sentence \(s\ = \{w_1, ..., w_n\}\) composed of \(n\) words, in \gls*{clm}, the maximum likelihood objective over \(s\) is:

\begin{equation}
\label{eq:clm}
    \mathcal{L}_{CLM} = \sum_{i=1}^{n} \log P(w_i|w_1,...,w_{i-1}).
\end{equation}

For our task, we are only interested in calculating the loss for the words from the profile sentence conditioned on the utterance.
Considering an utterance \(u = \{w_1^u,..., w_m^u\}\) and a profile sentence \(p = \{w_1^p,..., w_k^p\}\), we redefine the objective from Equation \ref{eq:clm}:

\begin{equation}
\label{eq:pgloss}
    \mathcal{L}_{PG} = \sum_{i=1}^{k} \log P(w_i^p|w_1^u,...,w_m^u,w_1^p,...,w_{i-1}^p).
\end{equation}

As seen in Equation \ref{eq:pgloss}, the loss is only calculated for the generation of the profile sentences.
In the model's input, we separate the utterance and profile sentences using a special token \texttt{<gen>} and, as it can exist more than one profile sentence, we add \texttt{<sep>} between the profile sentences.

\section{Experiments}

In this section, we evaluate Transformer decoders on the novel dataset and provide benchmark results for future research.
Additional experimental details are provided in Appendix \ref{appendix:generation-details}.

\subsection{Models}

\paragraph{GPT2} This model has achieved state-of-the-art results in various generation tasks \cite{GPT2}. We consider two different pre-trained versions that differ in size, the \texttt{gpt2-small} and \texttt{gpt2-medium} (details in Appendix \ref{appendix:models}).
\paragraph{DistilGPT2} This is a distilled version of GPT2, where it was trained under the supervision of GPT2 \cite{DISTIL}. The \texttt{distilgpt2} contains about half the size of GPT2 while still achieving competing performance in various \gls*{nlp} tasks.

\subsection{Metrics}

We follow common practices for text generation and report BLEU \cite{BLEU} and ROUGE \cite{ROUGE}, metrics that, respectively, measure the precision and recall between the generated and the golden text.
Additionally, we employ BERT Score \cite{BERT-SCORE}, an automatic metric that leverages BERT's \cite{BERT} contextual embeddings and matches words in candidate and golden sentences using cosine similarity.

\subsection{Results}
\label{sec:results}
In Table \ref{tab:gen-results}, we provide benchmark results for the \gls*{pgt}.
The models without fine-tuning fail to extract the correct profile information from the dialogue sentences, which is expected as their pre-training was on a large collection of unstructured text.
We observe that fine-tuning the models has a great impact on the overall performance, where \texttt{gpt2-small} achieves the higher scores in all metrics except BERTScore (for a minimal difference).
In Appendix \ref{appendix:gen-examples}, we provide some generated examples from the evaluated models.
The results obtained show promising advances in this task and we hope that this will introduce a new future research direction in this area.

\section{Related Work}

Recent research has focused on building personalized dialogue systems using profile information.
\citet{PERSONA-NEURAL} proposed a neural conversational model to capture background information and speaking style from interlocutors in dialogue.
\citet{PERSONACHAT} introduced a dataset composed of personas, which are essentially 3 to 5 profile sentences describing the speaker's profile.
\citet{PERSONALIZED-TRAITS} studied how to include profile information such as age, location, and interests by explicitly incorporating this knowledge into the sequence-to-sequence framework.

Few works have attempted to identify profile knowledge from conversational data.
\cite{DETECT-SPEAKER} introduced a framework for detecting the correct profile amongst a set of distractor profiles.
Nevertheless, the authors do not consider the correlation between utterances and profile sentences.
\cite{DATA-MANIPULATION-PERSONA} proposed a data manipulation method to construct distilled and diversified dialogue data containing profile information and leverage it into the dialogue generation task.

\section{Conclusion}

We propose the \gls*{pgt} and contribute with \gls*{pgd}, a dataset with more than 30 000 pairs of utterances and profile sentences built with the feedback of human annotators. In addition, we train state-of-the-art models and achieve promising results in the proposed task. We hope that this new line of research will help the task of personalizing dialogues, although the task of automatically extracting profiles from dialogues is valuable by itself. 

\section*{Acknowledgements}
This work was supported by national funds through Fundação para a Ciência e a Tecnologia (FCT) with reference UIDB/50021/2020 and grant 2022.10640.BD, by the project CMU-PT MAIA with reference 045909, as well as by the Recovery and Resilience Plan (RRP) and Next Generation EU European Funds through project C644865762-00000008 Accelerat.AI.





\bibliography{custom}

\begin{thebibliography}{19}
\expandafter\ifx\csname natexlab\endcsname\relax\def\natexlab#1{#1}\fi

\bibitem[{Cao et~al.(2022)Cao, Bi, Fang, Shi, and
  Tao}]{DATA-MANIPULATION-PERSONA}
Yu~Cao, Wei Bi, Meng Fang, Shuming Shi, and Dacheng Tao. 2022.
\newblock A model-agnostic data manipulation method for persona-based dialogue
  generation.
\newblock In \emph{Proceedings of the 60th Annual Meeting of the Association
  for Computational Linguistics (Volume 1: Long Papers)}, pages 7984--8002.

\bibitem[{Gu et~al.(2021)Gu, Ling, Wu, Liu, Chen, and Zhu}]{DETECT-SPEAKER}
Jia-Chen Gu, Zhenhua Ling, Yu~Wu, Quan Liu, Zhigang Chen, and Xiaodan Zhu.
  2021.
\newblock Detecting speaker personas from conversational texts.
\newblock In \emph{Proceedings of the 2021 Conference on Empirical Methods in
  Natural Language Processing}, pages 1126--1136.

\bibitem[{Hinton et~al.(2015)Hinton, Vinyals, and Dean}]{DISTIL}
Geoffrey~E. Hinton, Oriol Vinyals, and Jeffrey Dean. 2015.
\newblock Distilling the knowledge in a neural network.
\newblock \emph{ArXiv}, abs/1503.02531.

\bibitem[{Kenton and Toutanova(2019)}]{BERT}
Jacob Devlin Ming-Wei~Chang Kenton and Lee~Kristina Toutanova. 2019.
\newblock Bert: Pre-training of deep bidirectional transformers for language
  understanding.
\newblock In \emph{Proceedings of NAACL-HLT}, pages 4171--4186.

\bibitem[{Kingma and Ba(2014)}]{ADAM}
Diederik~P. Kingma and Jimmy Ba. 2014.
\newblock Adam: A method for stochastic optimization.
\newblock \emph{CoRR}, abs/1412.6980.

\bibitem[{Li et~al.(2016)Li, Galley, Brockett, Spithourakis, Gao, and
  Dolan}]{PERSONA-NEURAL}
Jiwei Li, Michel Galley, Chris Brockett, Georgios Spithourakis, Jianfeng Gao,
  and William~B Dolan. 2016.
\newblock A persona-based neural conversation model.
\newblock In \emph{Proceedings of the 54th Annual Meeting of the Association
  for Computational Linguistics (Volume 1: Long Papers)}, pages 994--1003.

\bibitem[{Lin and Hovy(2002)}]{ROUGE}
Chin-Yew Lin and Eduard Hovy. 2002.
\newblock Manual and automatic evaluation of summaries.
\newblock In \emph{Proceedings of the ACL-02 Workshop on Automatic
  Summarization}, pages 45--51.

\bibitem[{Liu et~al.(2019)Liu, Ott, Goyal, Du, Joshi, Chen, Levy, Lewis,
  Zettlemoyer, and Stoyanov}]{ROBERTA}
Yinhan Liu, Myle Ott, Naman Goyal, Jingfei Du, Mandar Joshi, Danqi Chen, Omer
  Levy, Mike Lewis, Luke Zettlemoyer, and Veselin Stoyanov. 2019.
\newblock \href {http://arxiv.org/abs/1907.11692} {Roberta: A robustly
  optimized bert pretraining approach}.
\newblock Cite arxiv:1907.11692.

\bibitem[{Papineni et~al.(2002)Papineni, Roukos, Ward, and Zhu}]{BLEU}
Kishore Papineni, Salim Roukos, Todd Ward, and Wei-Jing Zhu. 2002.
\newblock \href {https://doi.org/10.3115/1073083.1073135} {Bleu: A method for
  automatic evaluation of machine translation}.
\newblock In \emph{Proceedings of the 40th Annual Meeting on Association for
  Computational Linguistics}, ACL '02, page 311–318, USA. Association for
  Computational Linguistics.

\bibitem[{Radford et~al.(2019)Radford, Wu, Child, Luan, Amodei, Sutskever
  et~al.}]{GPT2}
Alec Radford, Jeffrey Wu, Rewon Child, David Luan, Dario Amodei, Ilya
  Sutskever, et~al. 2019.
\newblock Language models are unsupervised multitask learners.
\newblock \emph{OpenAI blog}, 1(8):9.

\bibitem[{Song et~al.(2021)Song, Wang, Zhang, Zhang, and Liu}]{BERTOVERBERT}
Haoyu Song, Yan Wang, Kaiyan Zhang, Weinan Zhang, and Ting Liu. 2021.
\newblock Bob: Bert over bert for training persona-based dialogue models from
  limited personalized data.
\newblock In \emph{Proceedings of the 59th Annual Meeting of the Association
  for Computational Linguistics and the 11th International Joint Conference on
  Natural Language Processing (Volume 1: Long Papers)}, pages 167--177.

\bibitem[{Song et~al.(2020)Song, Wang, Zhang, Liu, and
  Liu}]{GENERATE-DELETE-REWRITE}
Haoyu Song, Yan Wang, Weinan Zhang, Xiaojiang Liu, and Ting Liu. 2020.
\newblock Generate, delete and rewrite: A three-stage framework for improving
  persona consistency of dialogue generation.
\newblock In \emph{Proceedings of the 58th Annual Meeting of the Association
  for Computational Linguistics}, pages 5821--5831.

\bibitem[{Welleck et~al.(2019)Welleck, Weston, Szlam, and Cho}]{DNLI}
Sean Welleck, Jason Weston, Arthur Szlam, and Kyunghyun Cho. 2019.
\newblock Dialogue natural language inference.
\newblock In \emph{Proceedings of the 57th Annual Meeting of the Association
  for Computational Linguistics}, pages 3731--3741.

\bibitem[{Williams et~al.(2018)Williams, Nangia, and Bowman}]{MNLI}
Adina Williams, Nikita Nangia, and Samuel Bowman. 2018.
\newblock A broad-coverage challenge corpus for sentence understanding through
  inference.
\newblock In \emph{Proceedings of the 2018 Conference of the North American
  Chapter of the Association for Computational Linguistics: Human Language
  Technologies, Volume 1 (Long Papers)}, pages 1112--1122.

\bibitem[{Wu et~al.(2020)Wu, Li, Wang, Chen, Wong, Feng, Huang, and
  Wang}]{EXPLOIT-PERSONA}
Bowen Wu, MengYuan Li, Zongsheng Wang, Yifu Chen, Derek~F Wong, Qihang Feng,
  Junhong Huang, and Baoxun Wang. 2020.
\newblock Guiding variational response generator to exploit persona.
\newblock In \emph{Proceedings of the 58th Annual Meeting of the Association
  for Computational Linguistics}, pages 53--65.

\bibitem[{Xu et~al.(2022)Xu, Gou, Wu, Niu, Wu, Wang, and Wang}]{LONG-TERM}
Xinchao Xu, Zhibin Gou, Wenquan Wu, Zheng-Yu Niu, Hua Wu, Haifeng Wang, and
  Shihang Wang. 2022.
\newblock Long time no see! open-domain conversation with long-term persona
  memory.
\newblock In \emph{Findings of the Association for Computational Linguistics:
  ACL 2022}, pages 2639--2650.

\bibitem[{Zhang et~al.(2018)Zhang, Dinan, Urbanek, Szlam, Kiela, and
  Weston}]{PERSONACHAT}
Saizheng Zhang, Emily Dinan, Jack Urbanek, Arthur Szlam, Douwe Kiela, and Jason
  Weston. 2018.
\newblock \href {https://doi.org/10.18653/v1/P18-1205} {Personalizing dialogue
  agents: {I} have a dog, do you have pets too?}
\newblock In \emph{Proceedings of the 56th Annual Meeting of the Association
  for Computational Linguistics (Volume 1: Long Papers)}, pages 2204--2213,
  Melbourne, Australia. Association for Computational Linguistics.

\bibitem[{Zhang et~al.(2019)Zhang, Kishore, Wu, Weinberger, and
  Artzi}]{BERT-SCORE}
Tianyi Zhang, Varsha Kishore, Felix Wu, Kilian~Q Weinberger, and Yoav Artzi.
  2019.
\newblock Bertscore: Evaluating text generation with bert.
\newblock \emph{arXiv preprint arXiv:1904.09675}.

\bibitem[{Zheng et~al.(2019)Zheng, Chen, Huang, Liu, and
  Zhu}]{PERSONALIZED-TRAITS}
Yinhe Zheng, Guanyi Chen, Minlie Huang, Song Liu, and Xuan Zhu. 2019.
\newblock Personalized dialogue generation with diversified traits.
\newblock \emph{arXiv preprint arXiv:1901.09672}.

\end{thebibliography}
\bibliographystyle{acl_natbib}

\clearpage

\appendix

\section{Fine-Tuning RoBERTa}
\label{appendix:roberta-dnli}

We fine-tune a pre-trained \texttt{roberta-base}\footnote{\href{https://huggingface.co/roberta-base}{https://huggingface.co/roberta-base}} \cite{ROBERTA} with 12 layers, 768 hidden units, 12 attention heads, and 125M parameters on 1 NVIDIA GeForce RTX 3080 to minimize the cross entropy. We use Adam \cite{ADAM} optimizer with a learning rate of \(5e^-5\).
The batch size was 32, we train for 20 epochs and early stop after 5 epochs without an increase in the validation accuracy.


\section{Profile Generation}

\subsection{Experimental Details}
\label{appendix:generation-details}

We perform 5 runs for each model on 1 NVIDIA GeForce RTX 3080 using different seed values and calculate the average score for all metrics.
Models are trained to minimize the cross entropy using Adam \cite{ADAM} optimizer with a learning rate of \(5e^-5\).
For \texttt{gpt2-small} and \texttt{distilgpt2}, the batch size was 16 while for \texttt{gpt2-medium} the batch size was 4 with 4 gradient accumulation steps.
We train for 20 epochs with early stopping where the training is stopped after 5 epochs without a decrease in the validation loss.
We generate the profile sentences with a maximum length of 50 and perform greedy sampling, i.e., select the next word with the highest probability.
All experiments are implemented using the HuggingFace\footnote{\href{https://huggingface.co/}{https://huggingface.co/}} and PyTorch\footnote{\href{https://pytorch.org/}{https://pytorch.org/}} libraries.

\subsection{Models}
\label{appendix:models}
The \texttt{gpt2-small}\footnote{\href{https://huggingface.co/gpt2}{https://huggingface.co/gpt2}} version contains 12 layers, 768 hidden units, 12 attention heads, and 117M parameters and \texttt{gpt2-medium}\footnote{\href{https://huggingface.co/gpt2-medium}{https://huggingface.co/gpt2-medium}} includes 24 layers, 1024 hidden units, 16 attention heads, and 345M parameters.
The distilled version \texttt{distilgpt2}\footnote{\href{https://huggingface.co/distilgpt2}{https://huggingface.co/distilgpt2}} \cite{DISTIL} is smaller than GPT2, where it is composed of 6 layers, 768 hidden units, 12 attention heads, and 82M parameters.

\subsection{Generated Examples}
\label{appendix:gen-examples}

As discussed in Section \ref{sec:results}, the fine-tuned models show promising results in the generation of correct profile sentences.
An example for that is presented in Table \ref{tab:gen-results-1}.
Here, all models successfully extract the profile sentence, although we could argue that \textit{loving} and \textit{liking} are semantically different.

However, the evaluation scores also show that the systems are still far from always extracting the correct profiles.
In Table \ref{tab:gen-results-2}, we show an example where all models generated unrelated profile information and failed to recognize that the speaker dropped off high school.

\begin{table}[h]
\centering
\begin{tabular}{p{\columnwidth}}
\toprule
\textbf{Dialogue Utterance:} \\
sorry, i do not like music, i like reading mystery books. \\
\midrule
\textbf{Golden Profile Sentences:} \\
i enjoy reading mysteries.\\
\midrule
\textbf{\texttt{distilgpt2}:} \\
i love to read mystery novels.
\\
\textbf{\texttt{gpt2-small}:} \\
i love reading mysteries in my free time.
\\
\textbf{\texttt{gpt2-medium}:} \\
i read mystery novels.
\\ \bottomrule
\end{tabular}
\caption{Generated example \#1 from fine-tuned models.}
\label{tab:gen-results-1}
\end{table}

\begin{table}[h!]
\centering
\begin{tabular}{p{\columnwidth}}
\toprule
\textbf{Dialogue Utterance:} \\
dropping out of high school was a bad idea. the landlord just called. \\
\midrule
\textbf{Golden Profile Sentences:} \\
i dropped out of high school.\\
\midrule
\textbf{\texttt{distilgpt2}:} \\
i just got a job at the elementary school in new england.
\\
\textbf{\texttt{gpt2-small}:} \\
my parents got a new job.
\\
\textbf{\texttt{gpt2-medium}:} \\
i just graduated high school.
\\ \bottomrule
\end{tabular}
\caption{Generated example \#2 from fine-tuned models.}
\label{tab:gen-results-2}
\end{table}

\end{document}